%% file: root_v2.tex
\documentclass[letterpaper, 10 pt, conference]{ieeeconf}  

\IEEEoverridecommandlockouts                              

\usepackage{graphicx} 
\usepackage{epsfig} 
\usepackage{mathptmx} 
\usepackage{times} 
\usepackage{amsmath} 
\usepackage{amssymb}  
\usepackage{booktabs}
\usepackage{multirow}
\usepackage{caption}
\usepackage{subcaption}
\usepackage[table]{xcolor}
\usepackage{xcolor}
\usepackage{calc}
\usepackage{tikz}
\usepackage{pgfplots}
\pgfplotsset{
	/pgfplots/colormap={gray}{rgb255=(0,0,0) rgb255=(0,0,0)},
	layers/my layer set/.define layer set={
	background,
	main,
	foreground
	}{
	},
	set layers=my layer set,
		/pgfplots/xlabel near ticks/.style={
		/pgfplots/every axis x label/.style={
			at={(ticklabel cs:0.5)},anchor=near ticklabel
		}
	},
	/pgfplots/ylabel near ticks/.style={
		/pgfplots/every axis y label/.style={
			at={(ticklabel cs:0.5)},rotate=90,anchor=near ticklabel}
	},
}

\tikzstyle{input} = [rectangle, 
minimum width=1cm, minimum height=1cm, text centered, draw=black, 
fill=blue!30]
\tikzstyle{process} = [rectangle, minimum width=1cm, minimum 
height=1cm, text centered, draw=black]
\tikzstyle{result} = [rectangle, minimum width=1cm, minimum height=1cm, 
text centered, draw=black, fill=orange!30]
\tikzstyle{line} = [thick,-,>=stealth]
\tikzstyle{arrow} = [thick,->,>=stealth]

\usepackage[hidelinks]{hyperref}  

\graphicspath{{./graphics/}}

\definecolor{car}{RGB}{100, 150, 245}
\definecolor{bicycle}{RGB}{100, 230, 245}
\definecolor{motorcycle}{RGB}{30, 60, 150}
\definecolor{truck}{RGB}{180, 30, 80}
\definecolor{othervehicle}{RGB}{0, 0, 255}
\definecolor{person}{RGB}{255, 30, 30}
\definecolor{motorcyclist}{RGB}{150, 30, 90}
\definecolor{bicyclist}{RGB}{255, 40, 200}
\definecolor{road}{RGB}{255, 0, 255}
\definecolor{parking}{RGB}{255, 150, 255}
\definecolor{sidewalk}{RGB}{75, 0, 75}
\definecolor{otherground}{RGB}{175, 0, 75}
\definecolor{building}{RGB}{255, 200, 0}
\definecolor{fence}{RGB}{255, 120, 50}
\definecolor{vegetation}{RGB}{0, 175, 0}
\definecolor{trunk}{RGB}{135, 60, 0}
\definecolor{terrain}{RGB}{150, 240, 80}
\definecolor{pole}{RGB}{255, 240, 150}
\definecolor{trafficsign}{RGB}{255, 0, 0}
\definecolor{low}{RGB}{0, 35, 80}
\definecolor{high}{RGB}{255, 230, 50}
\newlength{\DepthReference}
\newlength{\HeightReference}
\newlength{\Width}%
\newcommand{\textb}[1]%
{%
	\settowidth{\Width}{#1}%
	\colorbox{#1}%
	{%
		\raisebox{-\DepthReference}%
		{%
			\parbox[b][\HeightReference+\DepthReference][c]{\Width}{\centering#1}%
		}%
	}%
}
\title{\LARGE \bf
On the calibration of underrepresented classes in \\LiDAR-based semantic 
segmentation}

\author{Mariella Dreissig$^{1, 2, *}$, Florian Piewak$^{1}$ and Joschka 
Boedecker$^{2}$%
\thanks{Acknowledgement: This publication was compiled as part of the research 
project "KI Delta Learning" (project number: 19A19013A) funded by the Federal 
Ministry for Economic Affairs and Energy (BMWi) based on a resolution of the 
German Bundestag.}%
\thanks{$^{1}$Mercedes-Benz AG, $^{2}$University of Freiburg, $^{*}$Primary 
contact: {\tt\small mariella.dreissig@mercedes-benz.com}}%
}

\begin{document}

\maketitle
\thispagestyle{empty}
\pagestyle{empty}

\begin{abstract}
	The calibration of deep learning-based perception models plays a crucial role in their reliability. Our work focuses on a class-wise evaluation of several model's confidence performance for LiDAR-based semantic segmentation with the aim of providing insights into the calibration of underrepresented classes. Those classes often include VRUs and are thus of particular interest for safety reasons. With the help of a metric based on sparsification curves we compare the calibration abilities of three semantic segmentation models with different architectural concepts, each in a in deterministic and a probabilistic version. By identifying and describing the dependency between the predictive performance of a class and the respective calibration quality we aim to facilitate the model selection and refinement for safety-critical applications.
\end{abstract}

\input{sections/introduction}
\input{sections/literature}
\input{sections/methods}
\input{sections/results}
\input{sections/discussion}
\input{sections/conclusion}

\bibliographystyle{IEEEtran}
\bibliography{IEEEabrv,export}

\end{document}

%% file: sections/introduction.tex
\section{Introduction}
\label{sec:introduction}
Environment perception allows an autonomous vehicle to detect and understand 
the behavior of other participants and enables it to adapt its own behavior 
accordingly. Deep learning methods took the performance in environment 
perception to a new level by evaluating large amounts of data gathered by 
various sensors with different modalities. Besides cameras and RADAR sensors, 
in the past years the LiDAR sensor gained relevance in the context of 
environment perception for autonomous vehicles due to the added value of highly precise depth information \cite{Li2020SPM}. 

Besides other perception tasks, semantic segmentation plays a crucial role in 
scene understanding for autonomous vehicles. The task of a semantic 
segmentation model is conducting a point-wise multi-class classification of
LiDAR point clouds \cite{Piewak2020PhD}, \cite{He2021arXiv}. Despite major 
advances in this field, the task of semantic segmentation comes with the 
challenge of handling severely imbalanced data \cite{Behley2021IJRR}, 
\cite{Triess2019Github}. This is due to the natural distribution of spaces and 
objects, i.e. in a traffic scene there always will be significantly more 
measurements of the road or buildings than of persons or bicyclists. 

These class imbalances have to be considered when developing and training a 
semantic segmentation model. While there have been approaches on overcoming 
this issue \cite{Paszke2016arXiv}, \cite{Piewak2019ECCV}, what remains unclear 
is the effect of class imbalance on the calibration of the model. In the 
context of autonomous driving, not only the detection of smaller instance 
classes is crucial but also having information about the reliability of those 
classifications. Ideally, the confidence should match the actual performance 
\cite{Guo2017ICML} and thus allow downstream tasks like sensor fusion or 
behavior planning to reliably interpret the models abilities. Thus, the effect 
of class imbalances on the calibration of a model is of particular interest 
regarding the safety of autonomous vehicles. 

This work focuses on the analysis of underderrepresented classes in terms of 
calibration in unmodified and probabilistic LiDAR-based semantic segmentation 
models. Our contributions can be summarized as follows: 
\begin{itemize}
	\item Design of a suitable calibration metric for semantic segmentation 
	models
	\item Analysis of model calibration given different confidence 
	measures
	\item Comparison of three semantic segmentation models and their 
	probabilistic versions in terms of class-wise calibration on LiDAR point 
	clouds
\end{itemize}

%% file: sections/literature.tex
\section{Related Work}
\label{sec:literature}
\subsection{Semantic Segmentation of LiDAR point clouds}
In the last years, various approaches on the semantic segmentation of point 
clouds have been proposed. Some operate in 3D space by utilizing voxels 
\cite{Huang2016ICPR}, \cite{Meng2018ICCV} or unordered point clouds 
\cite{Qi2017NIPS}, \cite{Li2018NIPS}, \cite{Yang2019CVPR}.
Other approaches project the point cloud into the 2D space in order use 
Convolutional Neural Networks developed for the camera-domain for the semantic 
segmentation of e.g. range view images \cite{Piewak2019ECCV, Chen2018ECCV, 
Wu2017ICRA, Xu2020ECCV, Aksoy2019IV}. 

Independently of the sensor modality, class imbalance is a problem which needs 
to be addressed in the semantic segmentation. Common ways to deal with it 
during training is to weight the loss function in favor of underrepresented 
classes \cite{Paszke2016arXiv}, \cite{Miloto2019IROS} or to construct an 
architecture which is able to overcome the issues imposed by smaller instances 
\cite{Piewak2019ECCV}, \cite{Cheng2021CVPR}. In terms of performance 
evaluation, the well-established mean Intersection over Union (mIoU) evaluation 
metric accounts for class imbalances and ensures that underrepresented classes 
are taken into account appropriately. 

\subsection{Calibration of Deep Learning Models}
In \cite{Guo2017ICML}, Guo et al. investigate the calibration of the 
traditional softmax probability and proved in a series of experiments its 
tendency towards overconfidence. This inspired a new field of 
research - uncertainty estimation in deep learning \cite{Arnez2020IJCAI}. 
Ensembling techniques like Monte-Carlo Dropout (MCD) \cite{Gal2016PhD} or Deep 
Ensembles \cite{Lakshminarayanan2017NIPS} are commonly used to approximate the 
unknown posterior of the model weights and are known to capture the model 
uncertainty well. Additionally, Kendall et al. \cite{Kendall2017NIPS} 
introduced a technique to use probabilistic logits to learn the data 
uncertainty directly from the input. 

Works like \cite{Mukhoti2021CVPR} and \cite{Gal2016PhD} argue that the softmax 
probabilities are less reliable even in models with additional uncertainty 
estimation techniques and propose using the entropy over the softmax 
probability distribution as uncertainty estimates. Yet, the raw entropy is 
not a probability and thus cannot be converted directly into a 
confidence measure. This results in the majority of works using the 
(calibrated) softmax as confidence measure instead \cite{Shen2021CVPR}, 
\cite{Gustafsson2020CVPR}, \cite{Cortinhal2020ISVC}, \cite{Cygert2021IJCNN}, 
\cite{Pearce2021arXiv}.

\subsection{Evaluation of Calibration Measures} \label{subsec:u_metric}
Guo et al. \cite{Guo2017ICML} proposed the Expected Calibration Error to 
assess the calibration of a given neural network, which is roughly speaking the 
correlation between the confidence and the accuracy of a model. While this 
produces an absolute measure and works well for tasks which actually use the 
accuracy as the performance metric, this approach tends to overestimate the 
performance for underrepresented classes. Nixon et al. \cite{Nixon2019arXiv} 
adapt this approach to multiclass settings, which in theory make it possible to 
calculate this metric for semantic segmentation tasks and weighting out the 
class-imbalances in the final score, but not solving the initial problem of the 
class-imabalances.

Mukhoti et al. \cite{Mukhoti2021CVPR} suggested a metric to capture particular 
parts of the calibration abilities of a semantic segmentation model. Their 
developed technique evaluates each frame patch-wise, which are labeled 
regarding their accuracy and uncertainty based on variable thresholds. This 
means, that this method requires some tuning which makes it difficult to use it 
for benchmarking different models.

Originally designed for the optical flow task, \cite{Ilg2018ECCV} proposed a 
calibration metric based on sparsification curves. It follows the same idea as 
\cite{Guo2017ICML}, that the confidence should coincide with the actual 
performance. Thus, they define the area under the sparsification error curve 
(AUSE) as a relative measure for a model's calibration. This has been applied 
to the task of semantic segmentation as well \cite{Shen2021CVPR}, 
\cite{Gustafsson2020CVPR} using the Brier score of the softmax probabilities to 
rate the predictive 
performance.

%% file: sections/methods.tex
\section{Methods}
\label{sec:methods}

\subsection{Semantic Segmentation Model and Training} \label{subsec:training}
We want to uncover which role the architecture choice plays in the calibration 
of underrepresented classes, thus we chose three very differently designed 
models for the semantic segmentation task: 1. a DeeplabV3+ model with a 
ResNet-50 encoder as backbone \cite{Chen2018ECCV}, 2. a SalsaNeXt  
\cite{Cortinhal2020ISVC} and 3. a LiLaNet, 
which was specifically designed for dealing with LiDAR point clouds 
\cite{Piewak2019ECCV}. While the former one uses residual blocks and intense 
pooling to reduce computational complexity, the latter one does not use pooling 
in order to conserve finer structures despite the low resolution. 

All models work on 2D data, thus the 3D point clouds are projected onto an 
image plane using the method published in \cite{Triess2020IV}. We use the 
SemanticKITTI dataset for autonomous vision tasks \cite{Behley2021IJRR} as a 
database. Both models are trained with a class-weighted cross-entropy loss 
based on \cite{Paszke2016arXiv} to account for the class imbalances and early 
stopping to avoid overfitting.

For all deterministic models we employ probabilistic versions by using MCD 
\cite{Gal2016PhD} (30 samples) and probabilistic logits \cite{Kendall2017NIPS} 
(5 samples) to model both aleatoric (data) and epistemic (model) uncertainty. 
For that, we add dropout layers (dropout rate of 0.5) in the middle part of the 
(encoder) model, following the findings of \cite{Kendall2017BMVC} and 
\cite{Mukhoti2021CVPR}.

\subsection{Confidence Measures}
As calibration measures we evaluate both the softmax and the entropy. 
It has to be noted that the softmax produces a probability distribution over 
the classes, summing up to $1$. The entropy in turn does not inherently exhibit 
the characteristics of a probability, but could be turned into one by 
normalizing with the theoretical maximum of $log(K)$. Yet, it produces only one 
confidence estimate conditioned on the given prediction instead of a 
distribution.

\subsection{Model Calibration}
Investigating the effects of underrepresented classes related to the 
calibration, we opted for a class-wise evaluation. Therefore, we need 
a metric which allows for a class-wise evaluation of any given confidence 
measure. At the same time, we want to decouple the calibration performance from 
the predictive performance as much as possible to avoid biases induced by the 
softmax.

\begin{figure}
	\centering
	\resizebox{0.9\columnwidth}{!}{%
	\begin{tikzpicture}[node distance=1.25cm and 0.5cm]
		\node (pred) [input, xshift=-1cm] {Prediction};
		\node (conf) [input, below of=pred] {Confidence};
		\node (perf) [input, xshift=2cm] {Performance};
		\node (sort) [process, below of=perf] {Sorting};
		\node (filter) [process, below of=perf, xshift=2cm] {Filtering};
		\node (point) [result, yshift=-0.75cm, xshift=7cm, align=center] 
		{Point on \\sparsification curve};
		\draw [line] (pred) -- (perf);
		\draw [line] (conf) -- (sort);
		\draw [line] (sort) -- (filter);
		\draw [arrow] (perf) -- (point);
		\draw [arrow] (filter) -- (point);
	\end{tikzpicture}}
		\caption{\textit{Elements of the AUSE metric.} Each point on the 
		sparsification curve is determined by three factos (marked in blue): 
		the prediction to determine the predictive performance on the $y$-axis 
		and the confidence, which will be implicitly entered on the $x$-axis 
		through ordering.}
	\label{fig:AUSE_contrib}
\end{figure}
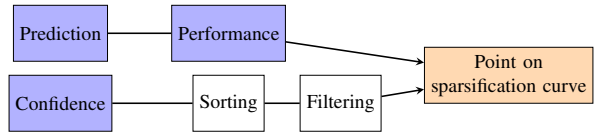

Sparsification plots have been previously used to evaluate confidence measures 
\cite{Wannenwetsch2017ICCV}. The idea is to evaluate the pointwise confidences 
by creating a ranking of the confidence values. The contributing factors of the 
sparsification curve are illustrated in Figure \ref{fig:AUSE_contrib}. The 
performance measure only depends on the prediction and is depicted on the 
$x$-axis. The confidence measure is depicted on the $y$-axis implicitly by 
ordinal sorting. The pixels with the lowest confidence are gradually removed 
and the performance on the remainder of points is evaluated. If the confidence 
measure actually reflects the true performance, the sparsification curve should 
monotonically increase.
Furthermore,  Ilg et al. \cite{Ilg2018ECCV} suggest a normalization based on 
the best possible ranking according to the ground truth labels to remove the 
dependence on the model performance, refered to as the oracle curve. The Area 
Under the Sparsification Error curve (AUSE) is defined as the area under the 
difference between the sparsification and the oracle curve. Intuitively, the 
closer the sparsification curve to the ground-truth-based oracle curve is, the 
smaller are the AUSE values and the better is the respective calibration. Thus, 
we expect the AUSE to be inversely correlated to the predictive performance in 
terms of IoU. 

The authors of \cite{Shen2021CVPR} and \cite{Gustafsson2020CVPR} have 
used the Brier score using the softmax probabilities to rate the predictive 
performance and the entropy as confidence (uncertainty) values. This imposes 
two issues with our setting:
\begin{enumerate}
	\item The Brier score requires the full probability 
	distribution over all classes, which some confidence measures other than 
	the softmax may not provide. Thus, it would not be possible to calculate 
	the Brier score based on e.g. the entropy.
	\item Using the entropy for the ranking ($x$-axis) but the softmax for the 
	Brier score on the $y$-axis introduces a mixing of both measures into the 
	metric (compare traces in Figure \ref{fig:AUSE_contrib}). As a result, we 
	would not know what influences the result most: the softmax probabilities 
	or the actual uncertainty estimation method. This makes it difficult to 
	draw conclusions from the results in order to further improve the model.
\end{enumerate} 
Due to these reasons, we use the IoU for each class as measure for the 
predictive performance. Additionally, we investigate both the softmax 
probability of the argmax prediction and the entropy over the softmax as 
confidence measures.

%% file: sections/results.tex
\section{Results}
\label{sec:results}
\subsection{Performance Evaluation} \label{subsec:pred_perf}
We calculate the IoU values to evaluate the predictive performance of all 
models and their variations. The results are listed in Table \ref{tab:IoU}. 
\begin{table}[]
	\centering
	\begin{tabular}{@{}lllllll@{}}
		\toprule
		& \multicolumn{2}{l}{DeeplabV3+} & \multicolumn{2}{l}{SalsaNeXt} & 
		\multicolumn{2}{l}{LiLaNet} \\ 
		class & deter. & prob. & deter. & prob. & deter. & prob. 
		\\ \midrule
		car & 0.86 & 0.85 & 0.91 & \textbf{0.93} & 0.90 & 0.91 \\
		bicycle & \cellcolor{gray!25}{0.00} & \cellcolor{gray!25}{0.01} & 0.04 
		& 0.05 & \textbf{0.11} & \textbf{0.11} \\
		motorcycle & \cellcolor{gray!25}{0.01} & \cellcolor{gray!25}{0.01} & 
		0.03 & 0.05 & \textbf{0.06} & \textbf{0.06} \\
		truck & \cellcolor{gray!25}{0.01} & \cellcolor{gray!25}{0.01} & 
		\cellcolor{gray!25}{\textbf{0.02}} & 		
		\cellcolor{gray!25}{\textbf{0.02}} & \cellcolor{gray!25}{\textbf{0.02}} 
		& \cellcolor{gray!25}{\textbf{0.02}} \\
		other-vehicle & 0.04 & 0.05 & 0.08 & 0.10 & 0.11 & \textbf{0.15} \\
		person & 0.05 & 0.07 & 0.17 & 0.19 & 0.20 & \textbf{0.23} \\
		bicyclist & 0.03 & 0.08 & 0.12 & 0.17 & 0.15 & \textbf{0.18} \\
		motorcyclist & \cellcolor{gray!25}{0.00} & \cellcolor{gray!25}{0.00} & 
		\cellcolor{gray!25}{0.00} & \cellcolor{gray!25}{0.00} & 
		\cellcolor{gray!25}{0.00} & \cellcolor{gray!25}{0.00} \\
		road & 0.90 & 0.88 & 0.93 & \textbf{0.94} & 0.92 & 0.92 \\
		parking & 0.06 & 0.06 & 0.10 & \textbf{0.11} & 0.08 & 0.09 \\
		sidewalk & 0.72 & 0.69 & 0.77 & \textbf{0.80} & 0.75 & 0.77 \\
		other-ground & \cellcolor{gray!25}{0.00} & \cellcolor{gray!25}{0.00} & 
		\cellcolor{gray!25}{0.00} & \cellcolor{gray!25}{0.00} & 
		\cellcolor{gray!25}{0.00} & \cellcolor{gray!25}{\textbf{0.01}} \\
		building & 0.73 & 0.72 & 0.85 & \textbf{0.86} & 0.81 & 0.82 \\
		fence & 0.11 & 0.13 & 0.22 & \textbf{0.24} & 0.20 & 0.22\\
		vegetation & 0.77 & 0.75 & 0.85 & 0.86 & 0.86 & \textbf{0.87} \\
		trunk & 0.35 & 0.33 & 0.46 & 0.50 & 0.48 & \textbf{0.51} \\
		terrain & 0.56 & 0.53 & 0.62 & 0.64 & 0.64 & \textbf{0.65} \\
		pole & 0.35 & 0.33 & 0.58 & \textbf{0.59} & 0.54 & 0.57 \\
		traffic-sign & 0.14 & 0.15 & 0.23 & 0.24 & 0.24 & \textbf{0.26} \\ 
		\midrule
		all & 0.39 & 0.41 & 0.52 & \textbf{0.56} & 0.47 & 0.51 \\ 
		\bottomrule
	\end{tabular}
	\caption{\textit{IoU values for all classes and mIoU in the validation 
	split of the SemanticKITTI dataset.} The values were calculated for both 
	the deterministic ("deter.") and the probabilistic ("prob.") version of 
	each model. The best performances are marked in bold.}
	\label{tab:IoU}
\end{table}
It is not surprising that in most cases the probabilistic versions performed 
better on the validation split than their deterministic counterparts. For the 
smaller instance classes, e.g. \textit{bicycle}, \textit{person} or 
\textit{traffic sign}, the LiLaNet outperforms the other models due to its 
architecture. Contrary, the well represented classes like \textit{car}, 
\textit{road} or \textit{building} are better learned by the SalsaNeXt model.

\subsection{Calibration Evaluation with AUSE} \label{subsec:calib_perf}
To gain insights about the calibration of the models, we calculate the AUSE 
across the full validation split. This ensures that even for the 
underrepresented classes enough pixels are evaluated. The mean values for both 
confidence measures over all frames and all classes can be seen in Table 
\ref{tab:overall_AUSE}.
\begin{table}[t]
	\centering
	\begin{tabular}{@{}lllllll@{}}
		\toprule
		& \multicolumn{2}{l}{DeeplabV3+} & \multicolumn{2}{l}{SalsaNeXt} & 
		\multicolumn{2}{l}{LiLaNet} \\
		& deter. & prob. & deter. & prob. & deter. & prob. 
		\\ \midrule
		softmax-based AUSE & 1.37 & 1.22 & 1.10 & \textbf{0.99} & 
		1.05 & 1.15 \\
		entropy-based AUSE & 1.36 & 1.22 & 1.10 & 1.00 & 1.08 & 
		1.17 
		\\ \bottomrule
	\end{tabular}
	\caption{\textit{Overall AUSE on the validation split of the 		
	SemanticKITTI dataset for all models.} The lower the value, the better 		
	is the model calibrated.}
	\label{tab:overall_AUSE}
\end{table}
The probabilistic SalsaNeXt achieves the best calibration, although the 
deterministic LiLaNet exhibits a similar AUSE value. Interestingly, in most 
cases the softmax probability actually outperformed the entropy in terms of 
calibration.

To gain deeper insights about the calibration of each class, we calculate the 
AUSE for all classes independently. Since the softmax performs slightly better 
than the entropy in terms of calibration, we focus further class-wise 
evaluations on the softmax. The results can be seen in \ref{tab:pc_AUSE}.

\begin{table}[t]
	\centering
	\begin{tabular}{@{}lllllll@{}}
		\toprule
		& \multicolumn{2}{l}{DeeplabV3+} & \multicolumn{2}{l}{SalsaNeXt} & 
		\multicolumn{2}{l}{LiLaNet} \\ 
		& deter. & probab. & deter. & prob. & deter. & prob. \\ \midrule
		car & 0.12 & 0.10 & 0.08 & 0.05 & 0.08 & 		
		\textbf{\underline{\textbf{0.05}}} \\
		bicycle & \cellcolor{gray!25}{\underline{0.53}} & 
		\cellcolor{gray!25}{1.32} & 2.13 & 
		2.12 & 2.15 & \textbf{1.99} \\
		motorcycle & \cellcolor{gray!25}{\underline{1.71}} & 
		\cellcolor{gray!25}{2.08} & 2.77 & 2.87 & \textbf{2.13} & 2.78 \\
		truck & \cellcolor{gray!25}{3.53} & \cellcolor{gray!25}{1.33} & 
		\cellcolor{gray!25}{0.75} & \cellcolor{gray!25}{\underline{0.15}} 
		& \cellcolor{gray!25}{1.33} & \cellcolor{gray!25}{1.72} \\
		other-vehicle & 2.69 & 2.75 & 3.00 & 2.89 & \underline{\textbf{2.19}} & 
		2.61 \\
		person & 2.48 & 2.06 & 0.92 & 0.82  & 0.81 & \textbf{\underline{0.76}} 
		\\
		bicyclist & 2.56 & 2.56 & 1.42 & \underline{\textbf{0.54}} & 1.30 & 
		1.06 \\
		motorcyclist & \cellcolor{gray!25}{\underline{0.00}} & 
		\cellcolor{gray!25}{\underline{0.00}} & 
		\cellcolor{gray!25}{\underline{0.00}} & 
		\cellcolor{gray!25}{\underline{0.00}} & 
		\cellcolor{gray!25}{0.04} & \cellcolor{gray!25}{0.52} \\
		road & 0.08 & 0.11 & 0.07 & \underline{\textbf{0.04}} & 0.05 & 0.05 \\
		parking & 3.32 & 2.75 & 3.25 & 2.74 & \textbf{\underline{2.31}} & 2.47 
		\\
		sidewalk & 0.58  & 0.70 & 0.48 & \textbf{\underline{0.43}} & 0.59 & 
		0.52 \\
		other-ground & \cellcolor{gray!25}{0.05} & 
		\cellcolor{gray!25}{\underline{0.01}} & \cellcolor{gray!25}{0.10} & 
		\cellcolor{gray!25}{0.30} & \cellcolor{gray!25}{0.74} & 
		\cellcolor{gray!25}{1.03} \\
		building & 0.31 & 0.23 & 0.11 & \textbf{\underline{0.10}} & 0.21 & 0.16 
		\\
		fence & 2.48 & 1.87 & \textbf{\underline{1.35}} & 1.91  & 2.13 & 2.19 \\
		vegetation & 0.50 & 0.51 & 0.20 & \textbf{\underline{0.17}} & 0.26 & 
		0.23 \\
		trunk & 1.35 & 1.13 & 1.21 & 0.83  & 0.89 & \textbf{\underline{0.77}} \\
		terrain & 0.54 & 0.70 & 0.52 & 0.47 & 0.43 & \textbf{\underline{0.41}} 
		\\
		pole & 1.53 & 1.41 & \underline{\textbf{1.01}} & 1.14 & 1.63 & 1.35 \\
		traffic-sign & 1.69 & 1.55 & 1.58 & 1.29 & 1.27 & 
		\textbf{\underline{1.18}}  \\ \bottomrule
		all (filtered) & 1.44 & 1.32 & 1.26 & \textbf{1.15} & \textbf{1.15} & 
		1.16 \\ \bottomrule
	\end{tabular}
	\caption{\textit{Per-class AUSE on the validation split of the 			
			SemanticKITTI dataset for all models.} The per-class AUSE is 
			calculated for softmax probability as confidence measure. The best 
			performances for each class are marked in bold, the underlined 
			values indicate the best performances after filtering for outlier 
			(marked in gray) in terms of predictive performance.}
	\label{tab:pc_AUSE}
\end{table}

\begin{figure}
	\centering
		\resizebox{0.85\columnwidth}{!}{%
		\begin{tikzpicture}
			\begin{axis}[
				enlargelimits=false,
				table/col sep=comma,
				xlabel={IoU},
				ylabel={AUSE},
				color={gray},
				style={ultra thick},
				ylabel near ticks,
				]
				\addplot+[
				only marks,
				scatter,
				mark=*,
				mark size=2pt]
				table
				{Mappe1.dat};
				\addplot[
				domain=0:1, 
				samples=100, 
				color=orange,
				on layer=foreground,
				style={ultra thick},
				]
				coordinates {(0.03,0)(0.03,3.5)};
			\end{axis}
	\end{tikzpicture}}
	\caption{\textit{Covariation of AUSE and IoU.} An inverse linear 
	relationship between the calibration and the predicitve performance would 
	be desireable. In that sense, the data points left of the orange line can 
	be seen as outliers due to a poor predictive performance.}
	\label{fig:AUSE_IoU}
\end{figure}
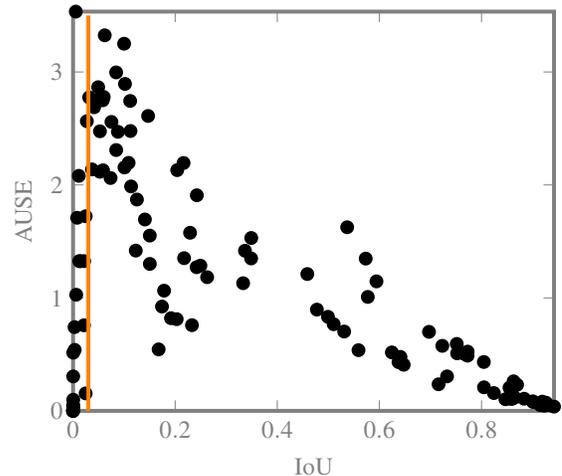

At first glance, no predominant tendency can be recognized from the calibration performance. When analyzing the AUSE in combination with the IoU values, the reason becomes obvious: some models were not able to learn some classes (e.g. \textit{motorcyclist}), resulting in an IoU of 0.0. Thus, for this class the oracle as well as the sparsification curve will constantly be 0.0, resulting in a perfect calibration score. This effect is depicted in Figure \ref{fig:AUSE_IoU}. We expect the relation between the predictive and the calibration performance to be rougly an inverse linear correlation. 

To gain some deeper insights into this phenomenon, we filter those cases (IoU $<$ 0.03, marked with a orange line) and re-evaluate the best calibration performances. After that, a similar pattern arises as in the IoU values in Table \ref{tab:IoU}: since the LiLaNet is better calibrated on smaller instance classes it's deterministic version exhibits a similar overall calibration performance as the probabilistic SalsaNeXt. It should be noted that the filtered classes are not neccessarily underrepresented classes but rather classes which are hard to learn from the available data.

%% file: sections/discussion.tex
\section{Discussion}
We demonstrated how a modification of the AUSE metric helps to analyze the 
calibration of a semantic segmentation model and to identify factors that 
contribute to the class-wise calibration. Following, we discuss our key 
findings based on Tables \ref{tab:IoU}, \ref{tab:overall_AUSE} and 
\ref{tab:pc_AUSE}.

\subsubsection{The SalsaNeXt performed best regarding the predictive and 
calibration performance, followed closely by the LiLaNet} The probabilistic 
version of the SalsaNeXt model exhibits the best calibration performance. Interestingly, the next best calibration performance is achieved by the deterministic LiLaNet. This might be due to its design, which inherently produces more calibrated softmax values. It illustrates that a probabilistic model with uncertainty estimation not necessarily exhibits a better calibration. It has to be noted that this finding is only supported by an evaluation on in-domain data. Contrary to the other two models, the DeeplabV3+ did not perform well on LiDAR data, suggesting that heavy pooling might not be beneficial when working with sparse and fine-grained structured data.

\subsubsection{The difference in calibration between softmax and entropy within a model is surprisingly small} We did not observe significant differences between the softmax and the entropy with respect to their calibration abilities. This indicates, that the softmax probabilities are able to capture uncertainty to some extend. It should also be noted that the calibration quality of the softmax influences the entropy, since it depends on the softmax probability distribution itself.

\subsubsection{Investigating the model calibration independently of the model performance exhibits several advantages} Our modified AUSE metric assesses the calibration performance independently of the chosen calibration metric and the model's predictive performance. Thus, it provides the basis for further investigations regarding the handling of underrepresented classes. Furthermore, it facilitates the model choice for any given task by enabling a custom mixing of the calibration with any predictive performance metric.

\subsubsection{The analysis of a mean calibration value over all classes might give misleading hints on the calibration performance} We observed the phenomenon of perfect calibration on unlearned classes which comes with decoupling the calibration performance from the predictive performance. That means, if a model is always wrong on a given class, the calibration will always be perfect. This effect can be avoided by constructing a model with a better performance on underrepresented classes or by filtering those classes.

%% file: sections/conclusion.tex
\section{Conclusion and Outlook}
In this paper we provided some insights about the role 
of class imbalance on the calibration of semantic segmentation models. We 
compared three models with different architectural characteristics, each in a 
deterministic and a probabilistic fashion. To evaluate the class-wise 
calibration performance, we modified sparsification-metric in order to decouple 
the predictive performance from the calibration. Furthermore, we gained some 
insights about the softmax compared to the entropy as confidence measures. Our 
key findings revealed that our metric is able to assess the calibration 
independently of the predictive performance, but in reality, the calibration 
and the predictive performance are influenced by each other. Furthermore, the 
calibration abilities depend on the structure of a model.

With this work we aim to promote research on the calibration on 
underrepresented classes and their effect on model performance and selection. 
Further research could include the evaluation of more model architectures, 
training strategies and uncertainty estimation methods. Additionally, it would
be of interest to refine the proposed metric related to the outlier filter 
which influeces the calibration performance. 